# Closed-loop AI achieves certifiable engineering design

Tianyi Yu[1,7], Chengxing Tao[1,7], Haoxuan Shen[1], Huiyang Li[1], Rugang Chen[1], Long Teng[1], Lilin Wang[1,2,3,4]*, Yan Li[1,5], Qingbin Chen[6], Chaogang Xu[6], Lizhong Wang[1,2,3,4]

1 State Key Laboratory of Ocean Sensing & Ocean College, Zhejiang University, Zhoushan, 316021, China

2 Zhejiang Province Key Laboratory of Offshore Civil Engineering and Materials & Ocean Academy, Zhejiang University, Zhoushan, 316021, PR China

3 Institute of Fundamental and Transdisciplinary Research, Zhejiang University, 310058, PR China

4 Hainan Institute of Zhejiang University, Sanya 572025, PR China

5 Goldwind Science & Technology Co., Ltd

6 State Power Investment Corporation Guangdong Electric Power Co., Ltd.

7 These authors contributed equally to this work

*Correspondence: lilin.wang@zju.edu.cn

# Abstract

Recent advances in agentic AI have demonstrated remarkable success in automating scientific discovery—generating research papers[1], writing expert-level code[2], proposing novel therapeutic candidates[3], and orchestrating autonomous experiments[4,5]. However, complex physical engineering design—where proposals must satisfy simultaneous constraints across fluid dynamics, solid mechanics, and structural stability—remains a notable gap. Here we introduce The AI Engineer, an agentic framework that orchestrates large language models(LLM)[6] with deterministic engineering backends in a closed loop: natural-language requirements are converted into design-domain geometry and mesh; topology is optimized with bi-directional evolutionary structural optimization (BESO)[7] coupled to the CalculiX solver[8]; member dimensions are refined with Particle Swarm Optimization (PSO) [9] coupled to Zwind software under real offshore aero-hydro-servo-elastic load cases[10].

To enable high-throughput autonomous exploration without incurring prohibitive per-candidate certification costs, The AI Engineer incorporates an Automated Reviewer as its primary internal termination gate. This module scores every candidate across five engineering dimensions—capacity, steel intensity, unit cost, constructability, and fatigue life—using piecewise linear functions calibrated against a reference fleet of 11 real floating wind projects [11-21]. The orchestrator terminates exploration only when a candidate achieves a composite score above a pre-validated threshold ($S \geq 85$, grade A) with no sub-dimensional score falling below 60. The threshold was empirically validated by submitting the single highest-scoring candidate to the China Classification Society (CCS) for full Approval in Principle (AIP)—a rigorous third-party review that the design passed, confirming that the automated gate is not an arbitrary heuristic but a faithful predictor of professional regulatory judgment. The AIP certificate therefore serves not as the system's daily objective function, but as the external gold-standard proof that the internal reviewer's logic aligns with human certification practice. The resulting design also outperforms the human-optimized "TuQiang" baseline, reducing steel consumption by 8.1% and unit capital cost by 8.1%, respectively, while satisfying all AIP criteria. This verification-closed regime—where every proposal is judged by deterministic physics and codified limit states before proceeding—distinguishes The AI Engineer from open-ended generative AI systems.
These results establish end-to-end automation of certifiable engineering design as a practical paradigm, with implications for rapid prototyping, design democratization, and certification-oriented AI workflows, while highlighting limits in detailed design and fabrication-hard constraints.

# Introduction

From the towering wind turbines that harvest renewable energy to the floating platforms that enable deep-water offshore power generation, the performance of engineered systems is fundamentally determined by their physical design. Unlike software, which can be iteratively debugged and patched, physical artifacts must satisfy simultaneous constraints across fluid dynamics, solid mechanics, and structural stability—all while remaining manufacturable, cost-effective, and reliable under uncertain operating conditions. For centuries, this design process has remained a quintessentially human endeavour, relying on the intuition, experience, and painstaking trial-and-error of expert engineers. Yet as physical systems grow increasingly complex—with tighter performance margins, stricter sustainability targets, and more tightly coupled multi-physics interactions—the limits of human-centric design have become starkly apparent. The question is no longer whether computational tools can assist physical design, but whether a fundamentally new paradigm—one that automates the entire design loop—can be achieved.

Consider the design of floating offshore wind turbines (FOWTs), one of the most urgent and demanding physical design problems of our era. To harvest wind energy in deep waters, a FOWT must balance conflicting objectives: maximizing energy capture while minimizing structural fatigue from unsteady aerodynamic and hydrodynamic loads, maintaining stability against violent wave-induced motions, and ensuring cost-competitive manufacturing and deployment. Each design iteration requires expensive high-fidelity finite-element and multi-body dynamics simulations, often taking hours and even days on computing clusters. The design space is high-dimensional and riddled with non-convex constraints. Consequently, the current state of practice remains heavily reliant on human experts, who iteratively propose candidate designs, interpret simulation outputs, hypothesize corrective modifications, and repeat—a laborious, time-consuming cycle that bottlenecks the entire engineering development pipeline (Fig. 1a).

Over the past decades, computational engineering has amassed an impressive arsenal of deterministic solvers—finite-element packages[22], boundary-element methods[23], and multi-body dynamics codes[24]—each capable of high-fidelity predictions for its assigned physics domain . Yet the engineering design cycle as a whole remains stubbornly human-intensive. The bottleneck is not the fidelity of individual solvers, but the orchestration across them. A typical floating-wind design iteration requires manual data translation between CAD modellers (geometry) [25], mesh generators (discretisation)[26], structural solvers (stress and stability)[27], and aero-hydro-servo-elastic simulators (coupled responses)[28]. When a continuum mesh degenerates, an element-set name dissociates from solver conventions, a geometric interference emerges after parametric updates, or a structural unity check[29] exceeds 1.0, the human expert must diagnose the root cause, adjust parameters or boundary conditions, restart the toolchain, and re-evaluate—a labour-intensive loop that fragments the design space and leaves vast regions of feasible topologies unexplored. Generative AI models trained on static design datasets cannot bridge this orchestration gap: they lack real-time access to the governing physics and are unaware of the numerical conventions, unit systems, and error codes enforced by each engineering tool. Reinforcement learning[30] and Bayesian optimization[31], while effective for tuning scalar objectives within a fixed simulation pipeline, are ill-suited to orchestrating heterogeneous software ecosystems and handling the non-differentiable, discrete failures—spanning geometric and meshing anomalies, numerical solver aborts, and physical limit-state violations—that dominate practical engineering workflows [32]. What is missing, therefore, is not a smarter surrogate or a faster solver, but an agentic orchestrator that monitors solver diagnostics, interprets numerical violations against codified limit states, and autonomously replans the next action—tightening mesh refinement, relaxing convergence criteria, or resetting the design domain—without demanding human attention at every failure. This is the gap we fill with The AI Engineer.

Here we introduce The AI Engineer, an end-to-end agentic framework that orchestrates LLM with deterministic engineering backends in a closed loop (Fig. 1b). Natural-language requirements are parsed into structured design specifications; the system then autonomously executes topology optimization (BESO coupled to CalculiX), parametric upscaling to target-class dimensions, size optimization (PSO coupled to Zwind), and automated drawing and report generation. Crucially, the orchestrator monitors every solver diagnostic, validates outputs against rule-augmented thresholds, and autonomously replans when mesh quality degrades, element-set semantics diverge, or structural unity margins are violated. The system terminates exploration not at local convergence alone, but when an internal Automated Reviewer—calibrated against real regulatory benchmarks—confirms that the candidate is certification-ready. Unlike open-ended generative AI systems that propose designs requiring external validation, The AI Engineer operates in a verification-closed regime: every proposal is judged by deterministic physics and codified limit states before proceeding, and the system terminates only when an internally validated certifiability gate—calibrated against real regulatory benchmarks—is passed. We demonstrate the framework on the design of a 20 MW FOWT foundation, a problem of central importance to China's 3060 Dual Carbon Goals[33] and one that exemplifies the multi-physics, multi-tool complexity that challenges conventional design methods  The AI-generated design was submitted to the China Classification Society (CCS) and passed formal Approval in Principle (AIP), confirming that the system can autonomously produce certifiable artifacts that rival—and in key metrics surpass—human-optimized baselines.

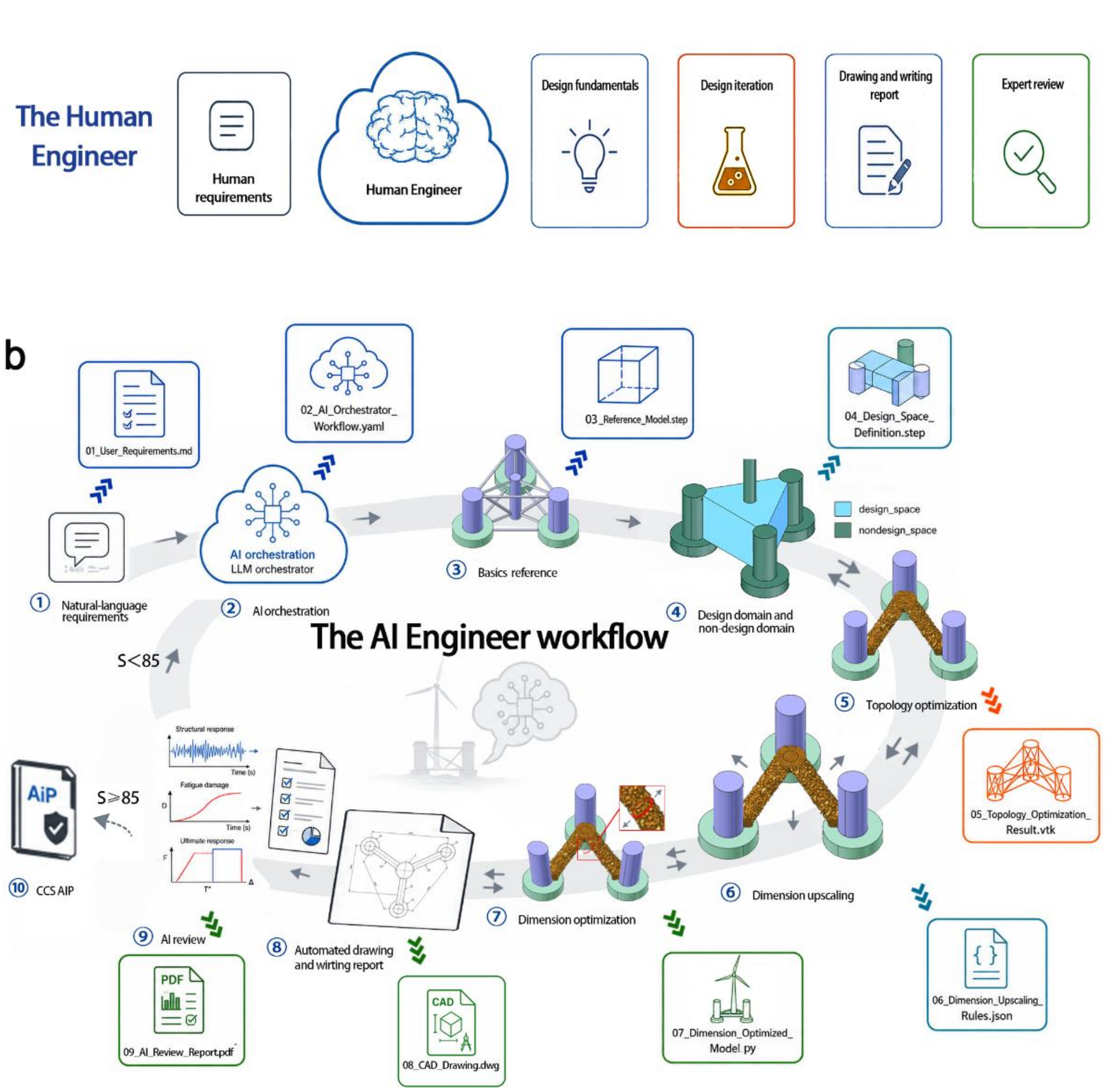

Fig. 1 | Comparison of the Human Engineer and The AI Engineer workflows. a, The Human Engineer workflow: defining fundamentals, determining primary dimensions based on a selected configuration, iterating designs, and passing third-party review. b, The AI Engineer workflow: the system operates in four autonomous phases (I–IV). Phase I formalizes natural-language requirements—owner's intentions, site-specific sea-state constraints, and classification-society provisions—into a structured specification. Phase II executes closed-loop optimization: it invokes the schema-validated toolchain; imports the reference model, such as OC4 reference model [34] for dimensional and layout priors; defines design and non-design domains in FreeCAD[25] and exports the optimizable mesh; performs compliance minimization with mass constraints via the CalculiX–BESO solver, recording the density field and topological evolution history; maps the topological skeleton to a target power-level configuration through parametric upscaling, generating a watertight STEP/FCStd solid model; performs PSO-based size optimization on the parameterized model; and evaluates the aero-hydro-servo-elastic coupled responses under operational and extreme load cases through Zwind simulations—with autonomous replanning triggered upon geometric anomalies, numerical solver aborts, or physical limit-state violations. Phase III produces engineering drawings and a structured design report from the optimized results. Phase IV invokes the Automated Reviewer as the primary internal termination gate: it evaluates the candidate across five engineering

dimensions—capacity, steel intensity, unit cost, constructability, and fatigue life—using piecewise linear scoring functions calibrated against a reference fleet of more than 11 real floating projects, delivering a composite score S (grade A-D); the orchestrator halts further exploration and selects the candidate for archiving only when $S \geq 85$ (grade A) with no sub-dimensional score below 60. This internal gate is empirically validated by a one-time external arrow: the single highest-scoring candidate is submitted to the China Classification Society (CCS) for full Approval in Principle (AIP)—a rigorous third-party review that the design passed, confirming that the Automated Reviewer threshold is a faithful predictor of professional regulatory judgment rather than an arbitrary heuristic. The AIP certificate thus sits outside the daily optimization loop (dashed external arrow) as the gold-standard calibration experiment, while the Automated Reviewer (solid internal arrow) governs routine termination decisions.

# Results

## AI-generated design outperforms the human-optimized baseline on key economic and technical metrics

We benchmarked the AI-generated foundation against the "TuQiang" baseline[21]—the world's first 20 MW floating wind demonstration unit, developed over two years by a consortium of more than 200 researchers and engineers from ten universities and industry partners. Figure 2a presents the three-dimensional configurations of both designs. The AI-generated candidate achieves substantial reductions in steel consumption (255.5 $t \cdot MW^{-1}$ versus 278 $t \cdot MW^{-1}$, an 8.1% saving) and unit capital cost (¥18197 $kW^{-1}$ versus ¥19,800 $kW^{-1}$, a 8.1% saving), while remaining well below the national demonstration target of 300 $t \cdot MW^{-1}$ [35](Fig. 2b-I to III). Critically, these economic gains are not achieved at the expense of structural integrity or dynamic safety. The fundamental natural frequency of the AI-generated design lies safely outside both the 1P and 3P excitation bands of the 20 MW turbine[36] (Fig. 2b-IV); platform motions under extreme sea states are comparable to or lower than the baseline (Fig. 2b-V); the maximum strength unity check remains below the allowable limit of 1.0 (Fig. 2b-VI); component fatigue damage accumulation stays within the 25-year design life with a Design Fatigue Factor of 3.0 (Fig. 2b-VII); and mooring line peak tensions are well below the breaking strength (Fig. 2b-VIII). These results establish that the AI system autonomously converged to a design that is simultaneously lighter, cheaper, and fully compliant with limit-state requirements.

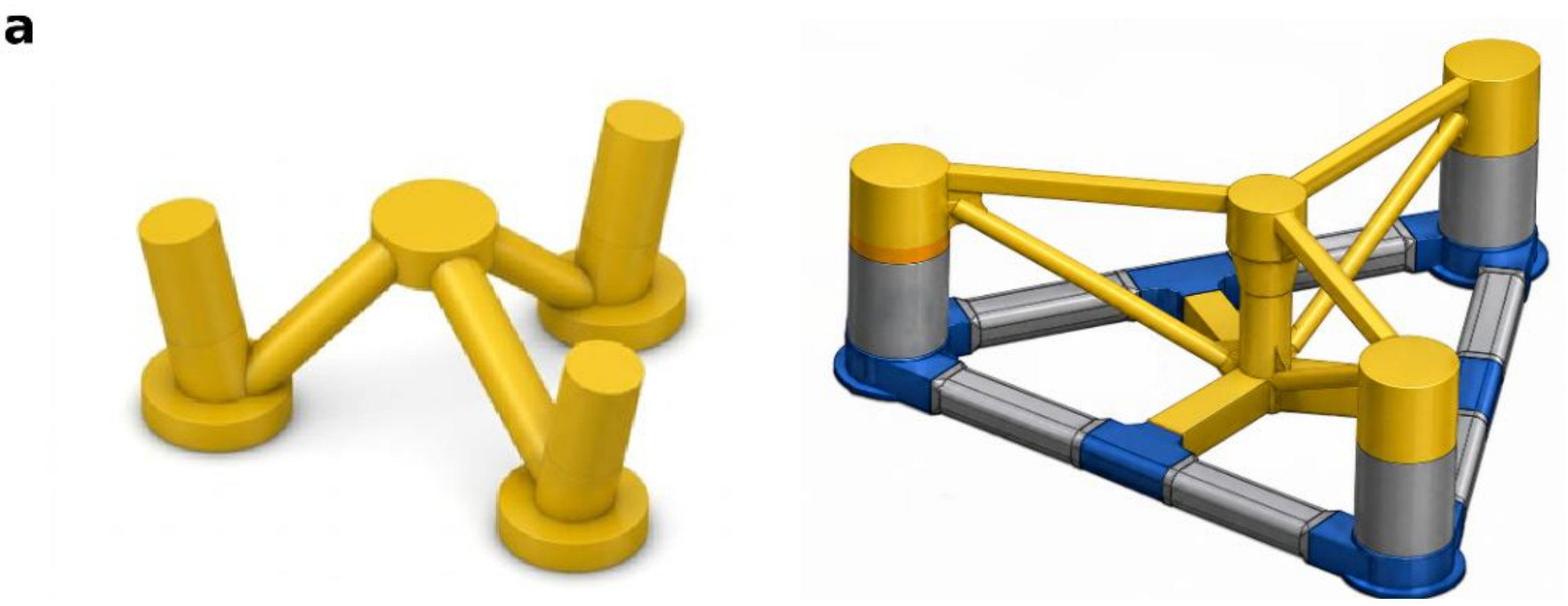

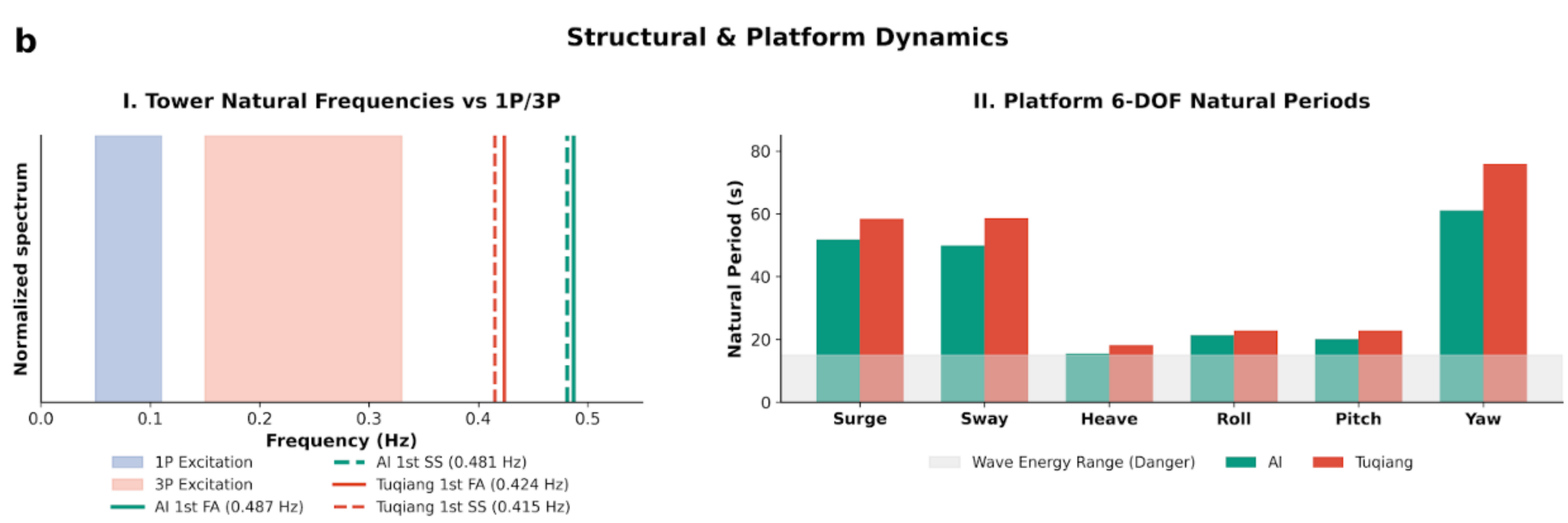
b
Structural & Platform Dynamics
I. Tower Natural Frequencies vs 1P/3P
Normalized spectrum
Frequency (Hz)
1P Excitation
3P Excitation
AI 1st FA (0.487 Hz)
AI 1st SS (0.481 Hz)
Tuqiang 1st FA (0.424 Hz)
Tuqiang 1st SS (0.415 Hz)
II. Platform 6-DOF Natural Periods
Natural Period (s)
Surge
Sway
Heave
Roll
Pitch
Yaw
Wave Energy Range (Danger)
AI
Tuqiang

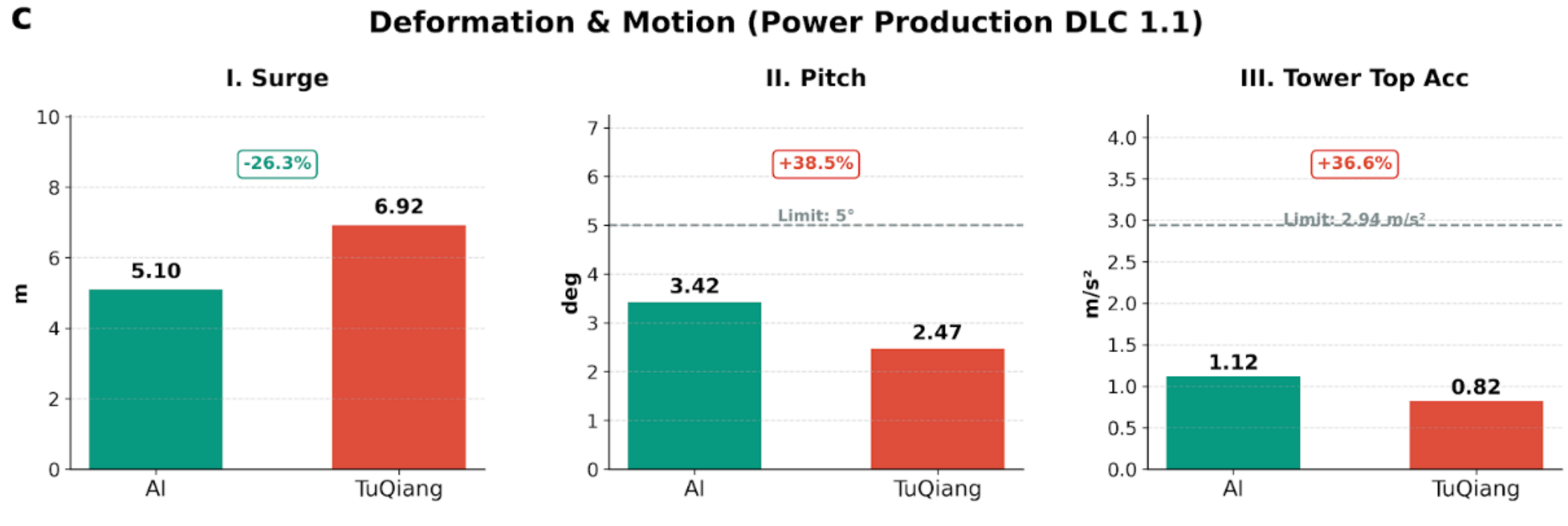
c
Deformation & Motion (Power Production DLC 1.1)
I. Surge
-26.3%
5.10
6.92
m
AI
TuQiang
II. Pitch
+38.5%
Limit: 5°
3.42
2.47
deg
III. Tower Top Acc
+36.6%
Limit: 2.94 m/s²
1.12
0.82
m/s²

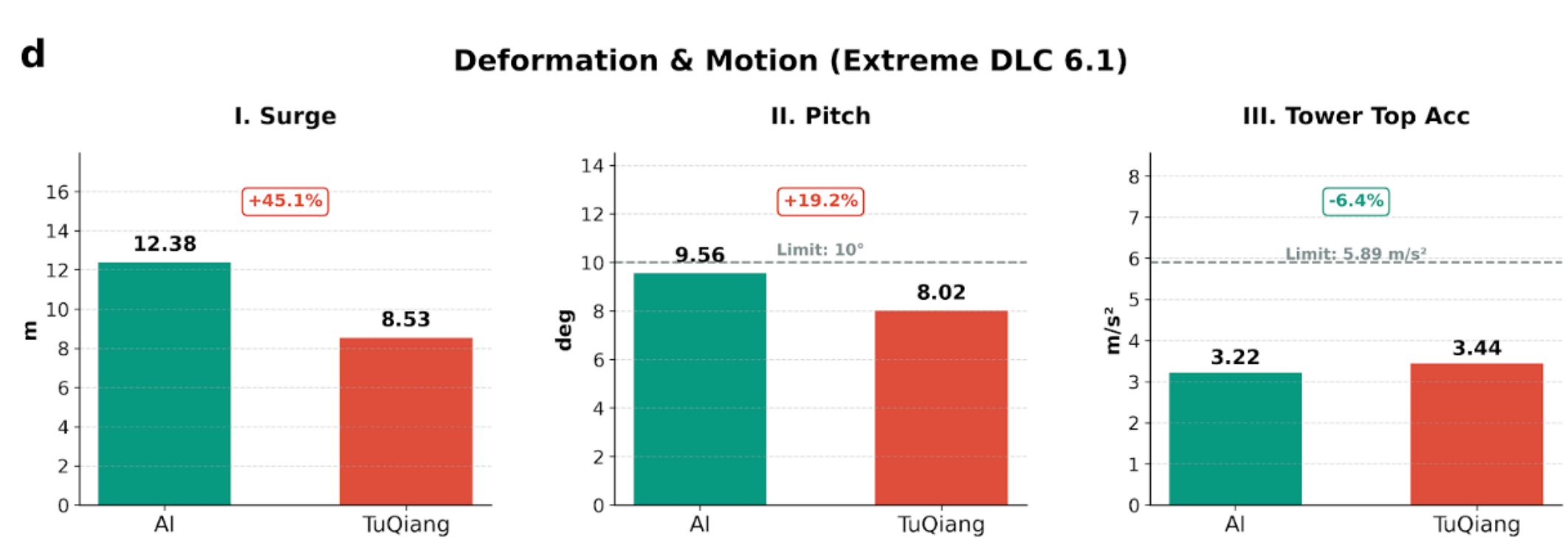
d
Deformation & Motion (Extreme DLC 6.1)
I. Surge
+45.1%
12.38
8.53
m
AI
TuQiang
II. Pitch
+19.2%
Limit: 10°
9.56
8.02
deg
III. Tower Top Acc
-6.4%
Limit: 5.89 m/s²
3.22
3.44
m/s²

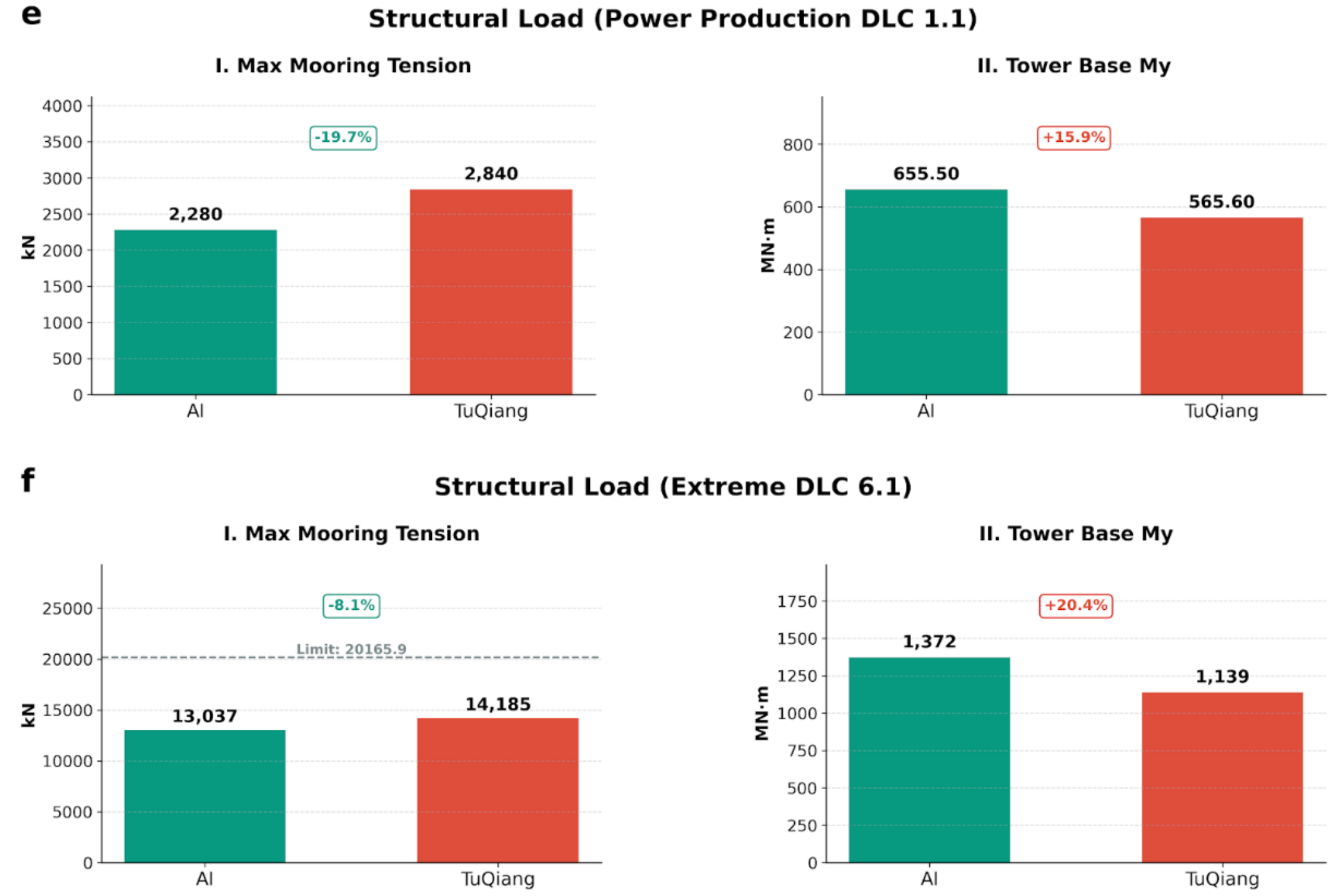

**Fig. 2 | Comparison between the AI-generated foundation and the "TuQiang" baseline. a**, 3D configuration of the AI-generated foundation (left) and the "TuQiang" reference (right). **b**, Structural and platform dynamics, showing the tower natural frequencies against the 1P/3P excitation bands (I) and the platform 6-DOF free-decay natural periods avoiding wave frequencies (II). **c**, Deformation and motion responses under the power production condition, comparing surge (I), pitch (II), and tower-top acceleration (III). **d**, Deformation and motion responses under the extreme sea state (DLC 6.1), evaluating surge (I), pitch (II), and tower-top acceleration (III). **e**, Structural strength and ultimate loads under the worst-case load condition, evaluating maximum mooring line tension (I) and tower-base maximum bending moment (II). Overall, the AI-generated design achieves reductions of 8.1% in steel mass, steel intensity, and unit capital cost—remaining well within the 300 $t \cdot MW^{-1}$ national demonstration target—while safely avoiding resonance with turbine excitation frequencies and enhancing structural limits.

## An Automated Reviewer for high-throughput internal screening is validated against regulatory benchmarks

Submitting every candidate generated during exploration to a classification society is economically and logistically infeasible. To enable high-throughput screening without sacrificing regulatory credibility, we equipped The AI Engineer with an Automated Reviewer—an internal scoring module that evaluates every candidate across five engineering dimensions (capacity, steel intensity, unit cost, constructability, and fatigue life) using piecewise linear functions calibrated against a reference fleet of more than 11 real floating wind projects. This fleet spans a capacity range from 2 MW to 20 MW and encompasses diverse floating topologies—from pioneering global prototypes (e.g.,

WindFloat[11, 13], Fukushima Shimpuu[12], Kincardine[14]) to state-of-the-art flagship demonstration platforms in China (e.g., Yinling[15], Fuyao[16], Guanlan[17], Mingyang Tiancheng[18], Gongxiang[19], Linghang[20], and TuQiang[21]). Detailed parameter specifications for the reference fleet are compiled in Methods and Supplementary Information.

To validate that this internal reviewer faithfully captures professional regulatory judgment, we ran a parallel regulatory review channel that independently scored the same reference projects against codified classification-society rules (DNV-ST-0437[37], DNVGL-OS-C301[38], and DNVGL-RP-0286[39]) rather than AI-optimized reference lines. Figure 3a shows the five-dimensional review profiles and the detailed score matrix for all 11 reference projects in the fleet. Furthermore, independent regulatory scoring of the fleet demonstrates that the AI and regulatory evaluation channels exhibit closely aligned trends across all dimensions. The quantitative agreement between the two channels is strong: the Spearman rank correlation coefficient[40] is 0.717 (Fig. 3b), the mean absolute deviation is 3.9 points, and 89% of projects achieve both scores above 60 with an absolute difference within 15 points—a threshold we define as high agreement. This validation confirms that the Automated Reviewer is not an arbitrary heuristic but a reliable surrogate for regulatory scrutiny, enabling the system to discard physically inferior or non-compliant candidates before they reach expensive limit-state verification or formal certification.

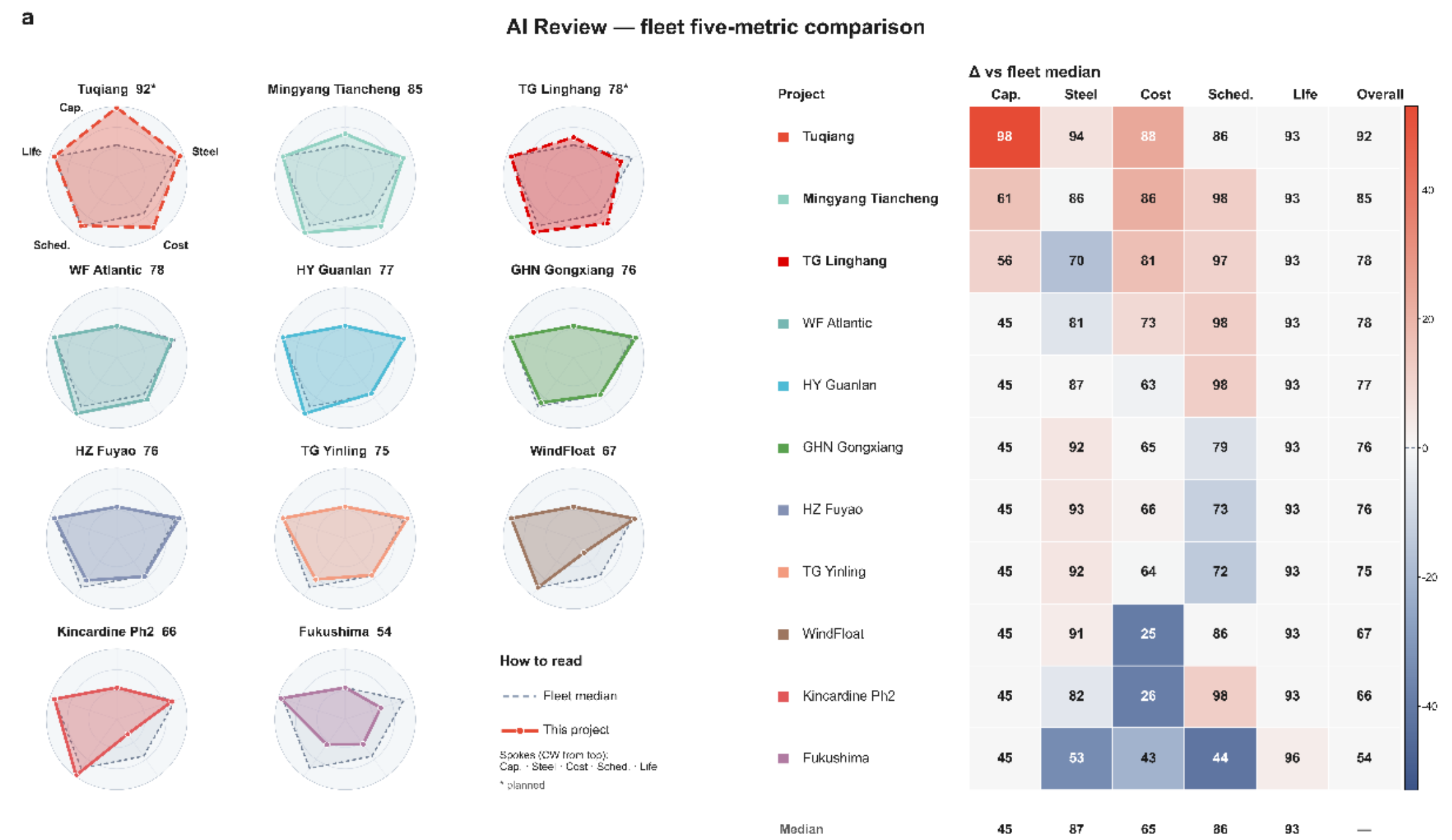

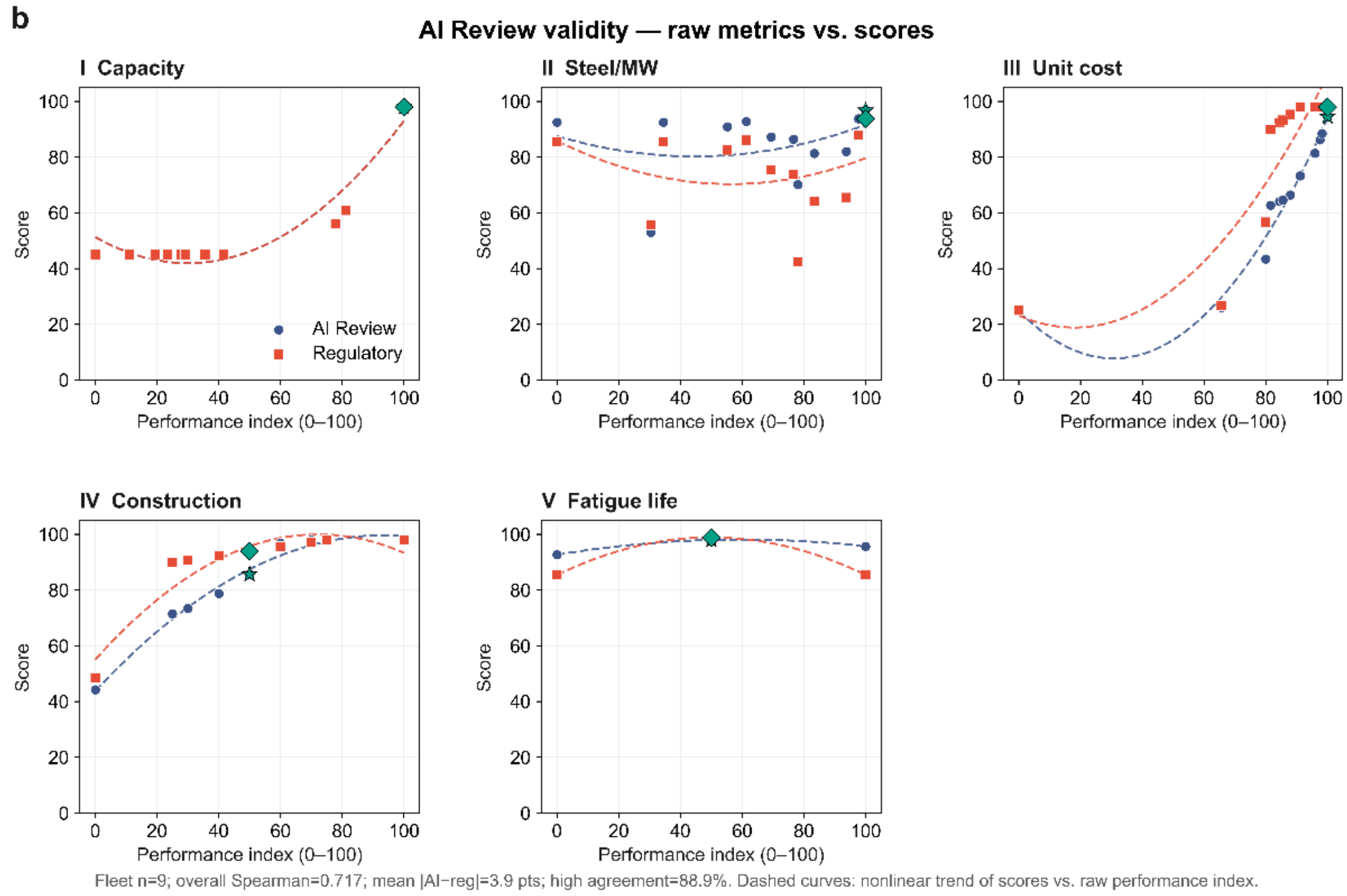

Fig. 3 | Validation of the Automated Reviewer framework against regulatory benchmarks and the reference project fleet. a, Five-dimensional review profiles and metric score matrix for the 11 reference projects. Individual radar charts (left) depict the performance relative to the fleet median across five engineering dimensions (capacity, steel intensity, unit cost, constructability, and fatigue life); the adjacent heatmap table (right) details the corresponding sub-metric scores and overall composite ratings. b, Correlation between Automated Reviewer scores and regulatory review scores across five dimensions, with each panel corresponding to one dimension: b-I, capacity; b-II, steel intensity; b-III, unit cost; b-IV, constructability; b-V, fatigue life. The horizontal axis shows the raw performance index from 0 to 100 normalized by fleet-internal percentile. Nonlinear trend curves were fitted with quadratic polynomials. The Spearman rank correlation coefficient is 0.717 and the mean absolute deviation is 3.9 points.

## External validation via formal third-party Approval in Principle (AIP)

To provide the ultimate gold-standard proof that the internal termination gate is calibrated correctly, we selected the single highest-scoring candidate generated by the system (composite score S = 91, grade A) and submitted its complete technical dossier—in full accordance with the CCS review checklist[41]—which comprised the design basis specification, general arrangement drawings, hull configuration drawings, capacity plan, intact and damaged stability calculation reports, loading condition calculation reports, hydrodynamic and motion analysis reports for the floating platform, global strength calculation and structural rule calculation documents, basic structural drawings, structural member classification drawings, mooring system calculation report, mooring layout drawings, and mooring line component drawings, together with the applicable codes list—to the China Classification Society (CCS) for formal Approval in Principle (AIP).

The AIP certification followed a four-stage workflow: preparation (defining the intended use, site conditions, and applicable standards), formal application (submission of all technical documentation), independent appraisal (CCS expert panel review, raising formal clarification requests and proposing optimizations), and final certification (issuance of the AIP certificate once all assessment items were resolved). During the independent appraisal stage, the CCS panel raised three rounds of formal clarification requests. The first round focused on the overall design rationale and fundamental static performance, verifying that the design basis and stability assessments were adequately justified. The second round addressed structural integrity and load-bearing capacity, with emphasis on strength verification and fatigue life assessment under combined environmental loads, as well as compliance with applicable code requirements. The third round concentrated on the station-keeping system and the design's robustness under extreme environmental conditions, including the verification of key component layouts and load paths. After three clarification rounds, all review items were closed and the AIP certificate was issued. Crucially, because The AI Engineer's orchestrator preserves a complete audit trail—including WebSocket logs, SHA-256 integrity manifests, and solver input decks—each clarification could be traced back to the exact computational step that produced the questioned output, substantially accelerating the response cycle[42]. After iterative exchanges and design refinements, the candidate successfully passed all review items and was awarded the CCS AIP certificate, confirming that the design meets the relevant code requirements at the conceptual level and possesses the technical basis for progressing to detailed design.

This external validation carries two important implications. First, it confirms that the Automated Reviewer's internal termination threshold ($S \geq 85$) is not a pre-calibrated heuristic but a genuinely predictive criterion: the selected candidate, which cleared this threshold, subsequently passed the full AIP process. Second, it establishes the AIP certificate not as the system's per-run objective function—which would be prohibitively expensive—but as a one-time gold-standard calibration experiment that validates the internal reviewer's trustworthiness. With this calibration in place, the system can subsequently explore thousands of new candidates at negligible marginal cost, relying on its internal Automated Reviewer for routine termination decisions while retaining confidence that the gate is aligned with third-party certification standards. Naturally, this AIP certification does not replace subsequent detailed design, fabrication inspection, or final type approval. Those remain outside the present scope and will be addressed in the next project phase. Nevertheless, the successful AIP submission provides the strongest possible empirical evidence that autonomous engineering design, grounded in deterministic solvers and regulated by an internally validated reviewer, can produce certifiable artifacts that rival—and in key metrics surpass—human-optimized baselines.

# Discussion

We have introduced The AI Engineer, an agentic framework that autonomously generates certifiable physical designs through a closed loop of deterministic simulation, rule-based threshold monitoring, and failure-driven replanning. The system produced a 20 MW floating offshore wind foundation that passed the China Classification Society's Approval in Principle (AIP) and outperformed the human-optimized "TuQiang" baseline—reducing steel intensity by 8.1% and unit capital cost by 8.1% while satisfying all coupled limit-state constraints[43]. More broadly, the framework establishes that agentic loops coupled with deterministic solvers can autonomously traverse the rigorous stages traditionally executed by expert consortia, from natural-language requirements to third-party-certifiable artifacts. The creative breakthrough lies in expanding the search space through BESO topology evolution and parameterized upscaling, strictly filtered by hard physics rather than unsupported text, and terminating not at local convergence but at an internally validated certifiability gate.

## Comparison with scientific discovery agents

How does an "AI Engineer" differ from an "AI Scientist"? Recent work has demonstrated multi-agent systems that automate scientific discovery—generating novel hypotheses[1,3], drafting expert-level empirical software[2], accelerating biomedical research[4], or orchestrating autonomous wet-lab experiments[5]. These systems excel at proposing plausible ideas that require external experimental validation; their success metrics are novelty, plausibility, and eventual reproducibility in the laboratory. The AI Engineer, by contrast, operates in what we term a verification-closed regime—meaning that every proposal is immediately evaluated against deterministic physics and codified limit states, leaving no room for unverifiable speculation. Every failure—whether a geometric meshing anomaly, numerical solver abort, or physical limit-state violation[29]—triggers an autonomous replanning action rather than a textual caveat. This has profound architectural consequences. Scientific discovery agents rely on language-based self-critique, retrieval-augmented reasoning, or offline training on static datasets; The AI Engineer's replanning decisions are numerically grounded—an element-set dissociation or a structural unity check of 1.05 generates a measurable solver diagnostic and a deterministic corrective response, not a probabilistic guess. Furthermore, the terminal condition of a discovery agent is often arbitrary (e.g., a fixed number of generations or a human-defined novelty threshold); the terminal condition of The AI Engineer is a formal third-party certificate obtained through the internal Automated Reviewer gate, empirically calibrated against real regulatory benchmarks. We believe this "verification-closed" paradigm defines a distinct class of AI agents for physical engineering, with accountability, auditability, and certifiability as first-class design constraints—constraints that are qualitatively different from those faced by scientific discovery systems.

## A two-stage validation cascade addresses the cost–credibility trade-off

A pragmatic objection to autonomous engineering design is that certification costs render the approach unscalable: submitting every candidate to a classification society is economically and logistically impossible. Our two-stage cascade directly resolves this tension.

The internal Automated Reviewer performs rapid, reproducible scoring across all candidates using piecewise linear functions calibrated against a reference fleet of more than 11 real projects. This scoring process is purely analytical—it evaluates the five engineering dimensions directly from the extracted metrics without invoking any finite-element or time-domain simulations—and completes in approximately 2–5 minutes per candidate on standard hardware. This low marginal cost enables the orchestrator to screen thousands of candidates before committing to a single AIP submission.

The orchestrator terminates exploration when a candidate achieves a composite score S ≥ 85 with no sub-dimensional score below 60. The external AIP submission serves as a one-time gold-standard calibration experiment, confirming that the internal threshold is not a heuristic but a faithful predictor of third-party certification. This cascade mirrors how human engineering teams operate: internal design reviews are cheap and frequent; external certification is reserved for the final selected concept. The strong correlation ($\rho_S$ = 0.717) and 89% high-agreement proportion between the Automated Reviewer and regulatory benchmarks demonstrate that the internal gate captures the essential logic of the human certification process. With this calibration in place, the system can subsequently explore thousands of new candidates at negligible marginal cost, retaining confidence that its termination decisions are aligned with third-party standards. The AIP certificate thus sits outside the daily optimization loop as the scientific instrument that proves the reviewer is trustworthy, not as a per-run objective function that would defeat the purpose of automation.

## Applicability to other engineering domains

Although we have demonstrated The AI Engineer on a 20 MW floating offshore wind foundation, the framework's architecture is fundamentally domain-agnostic. The orchestrator does not assume FOWT-specific physics; rather, it expects three enabling conditions that are met by a broad class of engineering problems. First, a deterministic solver (or suite of solvers) must be available to evaluate candidate designs against the relevant physics—structural finite-element analysis, boundary element method, multi-body dynamics, or coupled codes. The orchestrator interacts with these solvers through schema-validated input/output interfaces, requiring only that each solver returns quantifiable metrics and diagnostics. Second, the design problem must be governed by codified performance criteria that can be expressed as threshold-based constraints—spanning the four principal limit states (ULS, SLS, FLS, ALS), intact and damaged stability margins, dynamic resonance avoidance, and constructability limits. These criteria define the rule-augmented thresholds $\tau$ in the orchestrator's replanning logic. Third, a reference fleet of existing designs must be available to calibrate the Automated Reviewer's scoring functions, establishing a performance baseline against which candidate novelty and competitiveness can be judged.

Given these conditions, the framework is directly transferable to other marine and offshore structures (semi-submersible production platforms, jacket foundations, floating storage units), large-scale steel or concrete civil infrastructure (long-span bridges, high-rise structures, containment vessels), and aerospace components (airframe structures, turbine blades, landing gear) where high-fidelity solvers and certification standards are well established. For each new domain, the primary translation effort lies in redefining the geometry parameterization, the solver toolchain, and the regulatory threshold set $\boldsymbol{\tau}$—tasks that require domain expertise but no fundamental alteration of the orchestrator's decision-making or replanning logic.

The cost of such translation is modest relative to the benefit. In our FOWT implementation, configuring the schema-validated toolchain and calibrating the Automated Reviewer threshold required approximately three person-months of domain-specific engineering effort—a one-time investment that was then amortized over the autonomous exploration of thousands of candidates. For domains with existing digital design workflows and codified standards, we estimate a similar or lower integration effort. Conversely, domains lacking mature solvers (e.g., systems dominated by empirical correlations rather than first-principles physics) or without standardized certification pathways would not currently benefit from the framework. These represent boundary conditions for the current approach rather than fundamental limitations.

This domain-agnosticism, combined with the modular separation of the LLM-based orchestrator from the deterministic backends, suggests that The AI Engineer can serve as a general-purpose engineering exploration engine: a system that, given a solver, a set of constraints, and a performance baseline, autonomously searches for certifiable designs across problem domains. We therefore view the FOWT demonstration not as a domain-specific solution but as a proof of concept for a broader class of autonomous engineering systems—one in which human experts shift from operating individual design iterations to curating the solvers, standards, and objectives that the AI system orchestrates.

## Limitations and future directions

Although The AI Engineer autonomously generated a foundation that passed CCS AIP and outperformed the human-optimized baseline, several limitations remain before such systems can be deployed commercially.

First, although the system consistently generated multiple viable candidates that passed the internal screening thresholds, only the single highest-scoring branch was advanced to AIP submission; the others, while meeting all limit-state requirements, represented less optimal trade-offs across cost, constructability, or fatigue performance. The system's ability to consistently produce near-competitive designs—rather than its success rate in generating any feasible solution—must improve for routine industrial application. This likely requires tighter integration between the BESO topology module[7] and the downstream size optimizer, as well as more sophisticated replanning policies that can further elevate high-potential but suboptimal candidates into flagship-grade designs.

Second, while the system uses Zwind 2.5 for high-fidelity aero-hydro-servo-elastic time-domain simulation, resolving extreme localized hydrodynamics—breaking wave impacts, slamming forces, strongly nonlinear viscous drag, and complex failure mechanisms under wind-wave misalignment—ultimately requires physical wave-basin tests or prohibitive CFD campaigns. The current framework cannot autonomously orchestrate such experimental setups. Claims of typhoon survivability therefore require the same full-scale prototype validation planned for "TuQiang".

Third, fabrication constraints are currently prompt-approximated rather than CAD-CAM exact. The AI Engineer does not yet emit weld maps, non-destructive examination schedules, or yard lift plans. Detailed design and construction remain entirely human responsibilities. Incorporating manufacturing constraints as hard, differentiable penalties within the optimization loop is a priority for future work.

Fourth, the inherent stochasticity of LLM orchestration introduces epistemic variance. While the deterministic backends (CalculiX, Zwind) are reproducible given identical inputs, the orchestrator's high-level replanning decisions are influenced by the LLM's sampling behaviour. For regulatory submissions, we recommend multi-seed campaigns and strictly frozen tool schemas (with pinned dependency versions and deterministic solver flags) to guarantee reproducibility across runs.

Finally, "TuQiang" itself is a moving baseline. Continued national investment regularly compresses the frontier of cost and material efficiency, thus, claims of algorithmic superiority will therefore require periodic re-benchmarking against the latest human-optimized designs.

## Paradigm shift and ethical considerations

Despite these limitations, The AI Engineer establishes a broader point: agentic loops coupled with deterministic solvers can autonomously generate certifiable designs that rival—and in key metrics surpass—human-optimized baselines. Unlike text-only generative AI, engineering design imposes immediate numerical penalties for inconsistent units, degenerate meshes, or violated load limits. This makes transparent audit trails—WebSocket logs, SHA-256 integrity manifests, and versioned geometry fingerprints—not optional add-ons but critical infrastructure for establishing computational trust. Every stage of the design process is recorded under immutable content hashes, enabling third-party auditors to reconstruct the exact computational chain leading to a submitted design dossier. We believe this level of auditability is a prerequisite for any autonomous engineering system intended for safety-critical applications.

Ethically, certification societies must confront transparent AI authorship disclosure. Our protocol—AIP with pre-registered withdrawal clauses if certificate issuance would obscure AI attribution—offers a preliminary template, though global norms remain unsettled. This echoes concerns raised by recent autonomous discovery systems regarding the integrity of peer review and scientific attribution. For engineering certification, the stakes are even higher: if a design generated by an AI system is approved and later fails in service, the chain of accountability must be clear. We advocate that the human engineer responsible for submitting the design retains ultimate accountability, with the AI system treated as a computational collaborator rather than a legal author—a position consistent with the 'human-in-command' stance adopted by recent agentic systems in biomedical and scientific domains[44,45].

The framework optionally exposes an intervention interface at each phase boundary, allowing domain experts to inspect intermediate outputs and apply high-level adjustments—such as revising constraint priorities or freezing selected design subdomains—without breaking the closed-loop validation workflow. This interface serves scenarios where engineering judgement or project-specific requirements need to complement the optimization-driven search, preserving the human role as a high-level director rather than a low-level operator.

If developed responsibly, such agentic systems could broaden access to high-quality design iteration. They allow smaller engineering teams to explore thousands of physics-validated concepts before committing yard steel, while ensuring that accountable human engineers remain in charge of the final mile to type approval and offshore installation. The AI Engineer does not replace human engineers; rather, it amplifies their expertise through rigorous, physics-grounded automation. We envision that agentic frameworks of this kind will become standard collaborators in engineering practice—not as autonomous decision-makers, but as high-bandwidth explorers that dramatically expand the scope of systematic design exploration, leaving human judgment to navigate the trade-offs that no solver can encode.

# Methods

## The AI Engineer orchestrator

The AI Engineer orchestrates the complete lifecycle of physical system design autonomously. It couples a LLM reasoning layer with schema-validated deterministic engineering backends in a closed-loop architecture, translating natural-language specifications into certifiable numerical artifacts. Unlike conventional copilots that assist isolated tasks such as mesh generation or input-deck drafting, the system maintains continuity of state across sequential design phases: it decides what computation to invoke next, validates solver outputs against rule-augmented thresholds, and autonomously replans when geometric and meshing anomalies arise, numerical solvers abort, or physical limit-state margins are violated. Natural-language requirements $R$ are parsed into a structured job descriptor $\mathcal{J} = \{\text{phase}, \theta, \text{ retry_policy } \}$, where $\theta$ denotes phase-specific hyperparameters and "retry_policy" governs autonomous recovery actions.

The orchestrator may invoke only allowed engineering tools whose arguments are validated against formal schemas prior to execution, ensuring that language-model proposals cannot trigger arbitrary system commands or corrupt solver configurations. Each phase $p \in \{I, II, III, IV\}$ returns a standardized feedback tuple $\mathcal{F}_{p} = (\mathcal{L}_{p}, \mathcal{M}_{p}, \rho_{p}), \quad \rho_{p} \in \{0,1\}$: structured logs $\mathcal{L}_{p}$ (solver diagnostics, mesh inventories, convergence traces), quantitative metrics $\mathcal{M}_{p}$ (compliance, mass fraction, unity checks, hydrostatic margins), and a binary retry flag $\rho_{p}$. The closed-loop replanning rule is expressed as:

$$\text{If } \rho_p = 1 \text{ or } \exists m \in \mathcal{M}_{p}\text{: m violates } \tau,$$

$$\text{then } \theta \leftarrow \text{replan}(\theta, \mathcal{F}_{p}) \text{ and phase } p \text{ re-executes.} \qquad (1)$$

where $\tau$ denotes rule-augmented thresholds derived from classification-society design clauses.
The replan function $\text{replan}(\theta, \mathcal{F}_{p})$ maps the diagnostic signals in $\mathcal{L}_{p}$—solver exit codes, residual trajectories, or violated metric flags—to a predefined set of corrective actions. This diagnostic policy explicitly addresses three hierarchies of failures: (1) Geometric and meshing anomalies, where element-set dissociation triggers semantic remapping, and mesh quality degradation or geometric interference invokes localized remeshing and boundary adjustments; (2) Numerical solver aborts, where solver non-convergence is mitigated by tolerance relaxation or load-increment reduction, and transient instability by time-step adjustment; and (3) Physical limit-state violations, where strength, fatigue, kinematic, or resonance violations trigger parameter perturbation via the optimizer, and irrecoverable constraint conflicts invoke high-level design-domain backtracking.

This classification-driven approach ensures that replanning remains deterministic and auditable, enabling third-party reconstruction of recovery decisions without access to the LLM's internal states. Three representative cases are documented in Supplementary Table S.1. Deterministic backends span parametric solid modelling (FreeCAD[25]), continuum mesh generation (Gmsh), finite-element static and evolutionary solvers (CalculiX), gradient-free PSO for detailed size optimization and gradient-based SQP[46] for parametric upscaling refinement, and nonlinear time-history structural verifiers (Zwind 2.5). The orchestration layer sequences four autonomous phases: autonomous formulation of design prerequisites, autonomous configuration generation, autonomous size optimization, and automated evaluation and certification. A directed acyclic dependency is enforced, so no later phase may commence until the preceding phase satisfies its convergence criteria with $\rho_p = 0$.

Real-time observability is provided by streaming structured event logs to external observers, exposing tool invocations, parameter mutations, solver exit codes, and intermediate geometry fingerprints without requiring interactive terminal access. Artifacts at every stage are versioned under immutable content hashes, enabling third-party auditors to reconstruct the exact computational chain leading to a submitted design dossier. Tool-routing operates at low stochastic temperature to minimize epistemic variance in production certification workflows. Beyond deterministic pass/fail gates at each phase, an auxiliary evaluation layer scores candidate concepts against public benchmark platforms and multi-criteria rubrics spanning material efficiency, constructability, inspectability, and structural redundancy. This layer filters out linguistically plausible but physically inferior layouts before expensive limit-state verification.

The orchestrator terminates the generative loop and selects the final candidate for archiving only when three independent criteria are met simultaneously:

(1) Phase-level success: all phase retry flags remain inactive ($\rho_p = 0$ for all $p \in \{I, II, III, IV\}$), indicating that no autonomous replanning action is pending.

(2) Numerical convergence: the sizing optimization has reached its maximum number of evolutionary generations $n = 100$ with no further meaningful improvement observed and all coupled limit-state constraints (ULS, SLS, FLS, ALS) report zero maximum violation[47].

(3) Certification readiness: the Automated Reviewer composite score S meets the pre-calibrated threshold $S \geq 85$ (grade A), with no individual dimensional sub-score $s_i$ below 60. This composite score is defined as the weighted sum $S = \sum_{i=1}^{5} w_i \cdot s_i$ where the five dimensions (capacity, steel intensity, unit cost, constructability, fatigue life) carry default weights of 0.15, 0.30, 0.25, 0.15, and 0.15, respectively.

Condition (3) explicitly encodes certifiability as a non-negotiable component of the objective function, ensuring that the system does not converge to a lightweight structure that would fail regulatory scrutiny. The threshold S = 85 corresponds to the 90th percentile composite score among reference projects that have successfully passed independent regulatory reviews, and was empirically validated against the CCS AIP submission described in Results, where the selected candidate (S = 91) passed all review items. Once all three conditions hold, the orchestrator halts further exploration, archives the complete audit trail (WebSocket logs, SHA-256 integrity manifests, and solver input decks), and flags the selected candidate for external validation or detailed design. Representative failure-driven replanning cases—spanning geometric and meshing anomalies, numerical solver aborts, and physical limit-state violations—are documented in Supplementary Information, illustrating the orchestrator's diagnostic and recovery capabilities in practical engineering scenarios.

# Topology optimization

The system discovers the optimal load-transfer layout for the semi-submersible platform by minimizing global static compliance under a prescribed volume fraction constraint. The optimization, driven by a Bi-directional Evolutionary Structural Optimization (BESO) algorithm coupled with the CalculiX finite-element solver[8], is formulated as:

$$\text{Minimize} \quad C = \frac{1}{2}\mathbf{U}^{\mathbf{T}}\mathbf{K}\mathbf{U} \tag{2}$$

$$\text{Subject to} \quad V = \sum_{e=1}^{N_e} V_e \rho_e = m_{\text{goal}} V_0 \tag{3}$$

$$\rho_e \in \{x_{\min}, 1\} \tag{4}$$

where $\mathbf{U}$ and $\mathbf{K}$ are the global displacement vector and stiffness matrix, respectively. The binary design variable $\rho_e$ denotes the relative density of the e-th element; a soft-kill approach assigns a lower bound $x_{\min} = 10^{-3}$ to void elements to prevent stiffness-matrix singularities while effectively removing their structural contribution[48]. $V_e$ is the elemental volume, $V_0$ is the initial design domain volume, and the target mass retention fraction $m_{\text{goal}}$ governs the final material usage. Elemental sensitivity numbers—defined as the strain energy density—are smoothed using a spatial distance-based filter to suppress checkerboard patterns and mesh dependency. The algorithm simultaneously removes inefficient material from low-stress regions and reintroduces it to highly stressed void regions, controlled by prescribed addition and removal ratios that cap the permissible material change per iteration. Convergence is declared when the relative change in compliance falls below $10^{-3}$ over consecutive iterations and all auxiliary structural metrics (e.g., maximum and mean unity checks) remain within acceptable bounds. Full implementation details—including filter-radius evolution, adaptive mesh refinement strategies, and convergence histories—are provided in Supplementary Information.

## Parametric upscaling

The system parameterizes the conceptual geometry to accommodate the target rated capacity using an anisotropic, fixed-draft scaling paradigm[49]. Vertical dimensions are strictly preserved, while horizontal dimensions (column radius, inter-column spacing, and planar coordinates) are uniformly scaled by a factor $s$. The initial spatial approximation follows the rotor swept area scaling law, yielding a baseline horizontal multiplier $s_0 = \sqrt{P_{\text{target}}/P_{\text{base}}}$, where $P$ denotes the rated power.

This multiplier is subsequently refined by a correction variable $x$, such that $s = s_0 \cdot x$, through a preliminary sizing optimization that minimizes the specific steel consumption ($\text{t}\cdot\text{MW}^{-1}$). The structural mass is computed assuming hollow cylindrical sections of uniform wall thickness. The optimization enforces a static pitch angle $\theta \leq 5°$ under the maximum rated aerodynamic thrust:

$$\text{Minimize} \quad f(x) = \frac{m_{\text{struct}}(\text{x})}{P_{\text{target}}} \tag{5}$$

$$\text{Subject to} \;\; \theta(x) = \arctan\left(\frac{F_{\text{t}} \cdot H_{\text{hub}}}{K_{55}(\text{x})}\right) \leq 5° \tag{6}$$

where $m_{\text{struct}}$ is the total structural mass, $F_{\text{t}}$ is the rated aerodynamic thrust, and $H_{\text{hub}}$ is the hub height. The hydrostatic restoring stiffness in pitch, $K_{55}$, is analytically evaluated by integrating the waterplane area moment of inertia, displaced volume, and metacentric height. The constrained nonlinear optimization is solved using the SQP algorithm. This step establishes a hydrostatically viable and material-efficient envelope that serves as the macroscopic foundation for subsequent micro-dimensioning and integrity checks.

## Size optimization

The system employs Particle Swarm Optimization (PSO)[9] to yield the optimal design of the FOWT foundation. To transform the conceptual layout into specific engineering dimensions, size optimization is performed as the subsequent phase following topology optimization. Size optimization aims to minimize structural weight while satisfying functional constraints mandated by design codes. These constraints primarily encompass: (1) the mean platform tilt under operational conditions must not exceed 5°, and the maximum under operational conditions must not exceed 10°; (2) the utilization ratio (UC) for structural strength must remain below or equal to 1.0; (3) the fundamental natural frequency $f$ must lie outside the 1P and 3P frequency ranges to prevent resonance; and (4) cumulative fatigue damage D at critical welded joints must not exceed allowable design limits[50]. The size optimization formulation is therefore:

$$\text{Minimize } \text{W}(\mathbf{x}) = \sum_{\text{k}=1}^{\text{N}} \rho_{\text{steel}} L_k A_k(x_k) \tag{7}$$

$$\text{Subject to} \begin{cases} \theta_{\text{mean}}(\mathbf{x}) \leq 5° \\ \theta_{\text{max}}(\mathbf{x}) \leq 10° \\ \text{UC}(\mathbf{x}) \leq 1.0 \\ 1P \leq \text{f} \leq 3P \\ D(\mathbf{x}) \leq \frac{1}{\text{DFF}} \end{cases} \tag{8}$$

where $\rho_{\text{steel}}$ is the steel density, $\text{L}_\text{k}$ the length of the k-th member, $\text{A}_\text{k}(\text{x}_\text{k})$ the cross-sectional area,

DFF refers to the design fatigue factor.

The system refines tubular components (columns, braces, stiffeners) by solving a highly non-convex programming problem. The design vector $\mathbf{x}$ collects continuous cross-sectional parameters: shell thicknesses $t$ and outer diameters $D$. Because finite-element evaluations are computationally prohibitive within the heuristic loop, the system employs a gradient-free PSO algorithm, which iteratively updates candidate solutions using self-experience and swarm information.

To avoid repeated full finite-element analyses for each candidate, the system applies the Principle of Virtual Work (PVW) [51]to compute constraint sensitivities analytically. Virtual unit loads are applied at critical evaluation points, and the exact gradients of deflections and stresses with respect to x are obtained from a single linear static solution of the floating foundation under unit load conditions. These analytical sensitivities enable the PSO loop to evaluate three limit-state constraints concurrently without invoking the full time-domain solver at every particle evaluation, reducing computational overhead by orders of magnitude.

The sizing loop employs a fixed-generation termination strategy with a maximum of 100 generations, and the best solution is recorded every 10 generations.

# Zwind software

Zwind 2.5 resolves the nonlinear multi-physics coupling of utility-scale floating wind systems using Lagrange equations formulated for rigid-flexible coupled multibody systems. In generalized coordinates, the governing equation is:

$$\mathbf{M}(\mathbf{q})\ddot{\mathbf{q}} + \mathbf{C}(\mathbf{q}, \dot{\mathbf{q}})\dot{\mathbf{q}} + \mathbf{F}_{\text{int}}(\mathbf{q}) = \mathbf{F}_{\text{ext}} \quad (9)$$

where $\mathbf{M}(\mathbf{q})$ is the generalized mass matrix, $\mathbf{C}(\mathbf{q}, \dot{\mathbf{q}})$ the gyroscopic and Coriolis matrix, $\mathbf{F}_{\text{int}}(\mathbf{q})$ the internal structural force vector, and $\mathbf{F}_{\text{ext}}$ the external generalized force vector encompassing aerodynamics (BEM theory)[52], second-order hydrodynamics[53], mooring restoration, and control actions. Further details of the aero-hydro-servo-elastic coupling are provided in Supplementary Information.

# Automated Reviewer

The Automated Reviewer evaluates every candidate across five engineering dimensions—capacity, steel intensity, unit cost, constructability, and fatigue life—using piecewise linear scoring functions calibrated against a reference fleet of more than 11 real floating wind projects. For each dimension $i$, a sub-score $s_i \in [0,100]$ is computed. The composite score is:

$$S = \sum_{i=1}^{5} w_i \cdot s_i, \quad \sum_{i=1}^{5} w_i = 1 \quad (10)$$

with default weights [0.15, 0.30, 0.25, 0.15, 0.15] for the five dimensions respectively. The composite score maps to letter grades: A ($S \geq 85$), B ($70 \leq S < 85$), C ($60 \leq S < 70$), and D ($S < 60$). The $S \geq 85$ threshold corresponds to the 90th percentile among reference projects that have passed regulatory review, and was empirically validated by the single CCS AIP submission described in Results (the selected candidate scored S = 91 and passed all review items). The complete piecewise linear scoring functions for all five dimensions are provided in Supplementary Information.

To validate that the Automated Reviewer captures professional regulatory judgment, we ran a parallel regulatory review channel that independently scored the same reference projects against codified classification-society rules (DNV-ST-0437, DNVGL-OS-C301, DNVGL-RP-0286).

Agreement between the two channels was assessed through three statistics: the Spearman rank correlation coefficient ($\rho_S$ = 0.717), the mean absolute deviation (3.9 points), and the high-agreement proportion—defined as the fraction of projects for which both scores are ≥60 with an absolute difference ≤15—which was 89%. These validation results are shown in Fig. 3.

# AIP certification workflow

Approval in Principle (AIP) is a technical endorsement issued by CCS that independently verifies the feasibility of a design at the concept or preliminary design stage. It is not a substitute for final classification or detailed design approval, but rather confirms regulatory compliance and fundamental safety while explicitly requiring subsequent detailed design.

The AIP followed a four-stage workflow:

1. Preparation: defining the intended use, site conditions, applicable standards, and certification pathway.
2. Formal application: submission of technical documentation, sequentially comprising (1) principal dimensions, general arrangement drawings, and key system schematics; (2) hydrostatic stability assessments for intact and damaged conditions; (3) structural design calculations for ultimate and fatigue limit states; (4) mooring system analysis; and (5) a complete list of applicable governing codes[41, 54].
3. Independent appraisal: CCS expert panel review with formal clarification requests and optimization recommendations. During this stage, we responded with supplementary analyses and design revisions. The AI Engineer's complete audit trail—including WebSocket logs, SHA-256 integrity manifests, and solver input decks—enabled rapid tracing of each clarification to the exact computational step that produced the questioned output.

4. Final certification: the AIP certificate is issued once all assessment items were resolved.

The AIP certification does not supersede subsequent Type Approval or final classification. Detailed design, fabrication specifications, and offshore installation procedures remain outside the scope of this approval.

# Statistics and reproducibility

All flagship metrics are from single deterministic solver runs unless otherwise noted. The artifact bundle is accompanied by a SHA-256 integrity manifest for full reproducibility. All code and data are available as described below. Additional convergence histories, sensitivity analyses, and extended validation results are provided in Supplementary Information.

# Data and code availability

The high-fidelity simulation dataset and all flagship run bundles (INP, VTK, logs, beso_conf.py, Zwind reports) will be deposited in Zenodo with a DOI upon publication. Anonymized artifacts are available from the corresponding author on reasonable request. The AI Engineer orchestration platform is available at https://github.com/rainbowyuyu/ai_engineer under an open-source license. Third-party solvers (CalculiX 2.21, FreeCAD 0.21+, Gmsh 4.15, Zwind 2.5) follow their respective licenses. Full implementation details, including filter-radius evolution, adaptive mesh refinement strategies, and convergence histories, are provided in Supplementary Information.

## Acknowledgements

This work was supported by the Excellent Young Scientists Fund (Overseas), the National Key Research and Development Program of China (Grant No. 2023YFB4203302), the National Natural Science Foundation of China (52409144, 52238008), and "Pioneer" and "Leading Goose" R&D Program of Zhejiang (2025C01172).

## Author contributions

LL.W. and LZ.W. conceived the study, designed the research framework, developed the methodology, revised the manuscript, provided overall guidance and strong support for this study. LZ.W. is also Chief Scientist of TuQiang demonstration project. TY. Y. contributed to large language model part of the AI Engineer, whilst CX.T. contributed to optimization part of the AI Engineer. HX.S., HY.L., RG. C. and L.T. supported CX.T. to conduct optimization work. As the chief engineer of TuQiang foundation, Y.L. offered the design information of TuQiang demonstration project. As owner's representatives, QB. C and CG.X. provided the site information for the Tuqianghao project. All authors reviewed and approved the final manuscript.

## Declaration of interests

The authors of this paper are applying for national and international patents.

# Supplementary Information

This file contains Supplementary Methods detailing the implementation of the Bidirection Evolutionary Structural Optimization (BESO) solver, including softkill formulation, filter-radius evolution and adaptive mesh refinement strategies, as well as parametric upscaling sensitivity analysis, Automated Reviewer scoring functions with benchmark fleet data, and detailed mathematical derivations of the aero-hydro-servo-elastic multi-physics coupling for Zwind 2.5.

It also provides supplementary tables reporting the scoring results for the Approval in Principle candidate and the representative failure-driven replanning case studies encountered during the autonomous design campaign.

Supplementary figures include additional convergence histories, design-space exploration visualizations, and Automated Reviewer validation panels.

A supplementary algorithm presents the pseudocode for the closed-loop optimization procedure. Full implementation details, including configuration hyperparameters and audit-trail specifications, are provided to ensure reproducibility of the design.